# Advancements in Optimization: Adaptive Differential Evolution with Diversification Strategy

Sarit Maitra
Alliance Business school
Alliance University, Bengaluru, India
sarit.maitra@alliance.edu.in

**Abstract**

This study presents a population-based evolutionary optimization algorithm (Adaptive Differential Evolution with Diversification Strategies or ADEDS). The algorithm developed using the sinusoidal objective function and subsequently evaluated with a wide-ranging set of 22 benchmark functions, including Rosenbrock, Rastrigin, Ackley, and DeVilliersGlasser02, among others. The study employs single-objective optimization in a two-dimensional space and runs ADEDS on each of the benchmark functions with multiple iterations. In terms of convergence speed and solution quality, ADEDS consistently outperforms standard DE for a variety of optimization challenges, including functions with numerous local optima, plate-shaped, valley-shaped, stretched-shaped, and noisy functions. This effectiveness holds great promise for optimizing supply chain operations, driving cost reductions, and ultimately enhancing overall performance. The findings imply the importance of effective optimization strategy for improving supply chain efficiency, reducing costs, and enhancing overall performance.

## 1. Introduction

The goal of optimization is to maximize a system's desirable attributes while concurrently minimizing its unfavorable characteristics. While it is a well-known problem in applied machine learning and mathematical literature (Momin & Yang, 2013), it has a fair amount of share in different business domains, including supply chain analytics. Since its inception, the metaheuristic Differential Evolution (DE) algorithm (Storn & Price, 1995) has become a popular evolutionary algorithm to deal with optimization problems. Over the years, the applications of DE have been found in a wide spectrum of supply chain functions (e.g., Jauhar et al., 2017; Doolun et al., 2018; Yousefi & Tosarkani, 2022; Nimmy et al., 2022; Guo et al., 2023). The use of stochastic optimization techniques has increased over the past few years (e.g., Lan, 2020; Zakaria et al., 2020, etc.), and this has sparked significant interest within the research community regarding how optimization issues are handled. This motivated us to relook into the limitations encountered with the existing DE algorithms and the need for improved optimization techniques by introducing a new variant of DE.

The widespread application of DE can be found in diverse domains, including data mining, feature extraction, supply chain analytics, production scheduling, etc. A significant amount of research has already gone into improving this method and applying it to a range of practical issues (Opara & Arabas, 2019). In fact, since early 2010, researchers improved DE to improve its effectiveness and efficiency in handling various optimization challenges (Ahmad et al., 2022). This surge in popularity is attributed to DE's simple structure, rapid convergence, and robustness. However, DE's local search capabilities are often insufficient, making it prone to premature convergence and getting trapped in local optima (Wang et al., 2022; Liu et al., 2020). Additionally, when applied to high-dimensional optimization problems, DE's optimization accuracy tends to diminish (Cai et al., 2019; Liu et al., 2023). Maitra et al. (2023) have explored the scope of the existing DE approach and introduced a new algorithm to overcome the trap of local minima and enhance the performance of the algorithm. However, their algorithm was not validated with multiple benchmark test functions. Several variants of DE algorithms were introduced over the years to overcome the limitations of DE (e.g., Chakraborty, 2008; Mallipeddi, 2011; Draa et al., 2015; Deng et al., 2021; Xu et al., 2021). These variations have attempted to address problems like early convergence and robustness to various kinds of problems. Their advancement highlights the ongoing research efforts to enhance and broaden the capabilities of DE and associated optimization methods. Even with the improvement, considering its popularity, there is a growing need for in-depth research addressing these limitations and seeking potential enhancements to further elevate DE's effectiveness as an optimization tool.

This study introduces the Adaptive Differential Evolution with Diversification Strategies (ADEDS) algorithm, which improves on the limitations of the classic DE algorithm. ADEDS overcomes limitations in parameter tuning, local search capabilities, and high-dimensional optimization. Its adaptive system controls parameters, reducing human inference load. ADEDS's adaptability broadens its applicability to a wider range of optimization problems. It addresses early convergence issues by using dynamic diversification measures, allowing individuals to explore various alternatives without avoiding early convergence. This feature enhances ADEDS's ability to manage complex landscapes with multiple local optima. Therefore, the primary contribution of this work is the introduction and validation of the ADEDS algorithm, which addresses limitations in traditional DE for optimization tasks. The connection to supply chain analytics is made in this work to underscore the real-world implications and significance of the algorithm's effectiveness in optimization tasks. It is no secret that today's supply chain managers use technology to coordinate activities throughout the supply chain and provide important insights into performance.

## 2. ADEDS - Adaptive Differential Evolution with Diversification Strategies

To address the issues of premature convergence and lack of diversity in Differential Evolution variants, we propose the ADEDS algorithm. It is an evolutionary optimization algorithm that falls under the category of population-based stochastic optimization methods. We compare ADEDS with traditional DE, which is valuable in supply chain analytics, to identify advanced optimization techniques that can outperform traditional methods, thereby enhancing supply chain efficiency.

The purpose of this analysis is to discover the best solution for the sinusoidal function as displayed in Fig. 1, which is difficult for typical Differential Evolution (DE) techniques to solve.

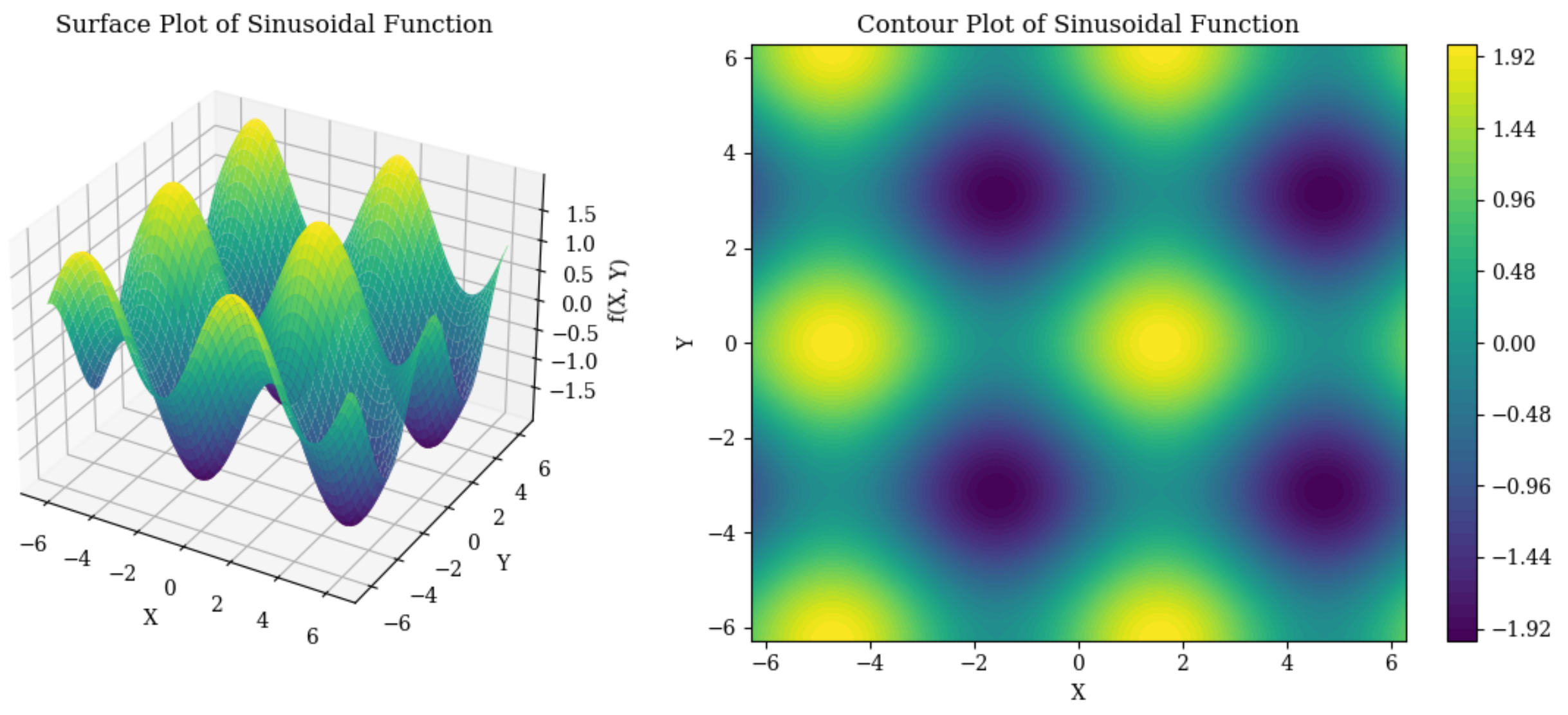

Fig. 1. Non-Convex Sinusoidal Function Landscape

- Initialization: $Population = [x_1, x_2, \ldots., x_{population_size}]$, each $x_1$is a vector in the search space with n dimensions. Table 1 displays the pseudo code of population initiation.

Table 1. Pseudo code for population initiation

```
Function initialize_population(population_size, bounds):
   population = List ()

   For i = 1 to population_size:
      individual = []

      For j = 1 to length(bounds):
         # Generate a random value within the specified bounds for dimension 'j'
         random_value = RandomUniform(bounds[j]. low, bounds[j]. high)
         Add random_value to individual

      Add individual to population

   Return population # Return the populated list of individuals.
End Function
```

- Adaptive mutation rate: $F = initial_mutation_rate * \left(1 - \frac{generation}{max_generation}\right)$. Table 2 displays the pseudo code of adaptive mutation.

Table 2. Pseudo code for adaptive mutation

```
Function adaptive_mutation_rate (generation, max_generations, initial_mutation_rate):
   # adaptive mutation rate based on generation and max_generations
   mutation_rate = initial_mutation_rate * (1.0 - generation / max_generations)

   Return mutation_rate
End Function
```

- Adaptive crossover rate: $CR = initial_crossover_rate * \frac{generation}{max_generation}$. Table 3 displays the pseudo code of adaptive crossover.

Table 3. Pseudo code for adaptive crossover rate

```
Function adaptive_crossover_rate(generation, max_generations, initial_crossover_rate):
   # adaptive crossover rate based on generation and max_generations
   crossover_rate = initial_crossover_rate * (generation / max_generations)

   Return crossover_rate
End Function
```

- Mutation strategy: Calculate the trial solution $v_i$ for each candidate solution $x_i$ using mutation strategies.

$$v_i = x_{r1} + F * (x_{r2} - x_{r3})$$

$r1, r2, r3$ are distinct random indices representing different solutions in the population, F is the adaptive mutation factor.

- Crossover operation: Combine the trial solution $v_i$ with the original solution $x_i$ to create a new candidate solution $u_i$. Table 4 displays pseudo code combining mutation strategy and crossover operation.

$$u_i[j] = \begin{cases} v_i[j] & if\ \ j = random(0,1) < CR \\ x_i[j] & otherwise \end{cases}$$

Table 4. Pseudo code Mutation strategy & crossover operation: F is a scaling factor for mutation. It is dynamically adjusted based on the generation progress using the adaptive_mutation_rate function.

```
For i from 0 to population_size - 1:
    individual = population[i] # Get the current individual

    # adaptive mutation rate and crossover rate
    F = adaptive_mutation_rate(generation, max_generations)
    CR = adaptive_crossover_rate(generation, max_generations)

    neighbors = list(range(population_size))

    # Randomly select two distinct neighbors
    neighbor1 = population[select_random_element(neighbors)]
    neighbor2 = population [select_random_element(neighbors, exclude=neighbor1)]

    # trial solution using DE mutation strategy
    trial_solution = individual + F * (neighbor1 - individual) + F * (neighbor2 - individual)

    # fitness of the trial solution
    trial_fitness = objective_function(trial_solution)

    # Update the population with the trial solution if it's better
    if trial_fitness < fitness[i]:
        population[i] = trial_solution
        fitness[i] = trial_fitness
End For
```

- Local Search: After generating a trial solution $v_i$, apply a local optimization algorithm to refine $v_i$ and potentially replace $x_i$ with a better solution. Table 5 displays pseudo code of local search.

$$x_i = local_optimization(v_i)$$

Table 5. Pseudo code for local search

```
Function local_search(solution):
    result = minimize (objective_function, solution, method='L-BFGS-B', bounds=bounds)

    # Return the optimized solution
    Return result.x
End Function
```

- Select candidate solutions for the next generation based on their fitness values. Solutions with lower fitness values are favored.

$$Population = select_new_population(population)$$

- Monitor the convergence of the optimization process. In this implementation, the algorithm stops when either of the following conditions is met:
    - It reaches the maximum number of generations specified by max_generations.
    - Fitness values have stagnated for stagnation_limit consecutive generations.
    - These criteria help prevent the algorithm from running indefinitely and provide a mechanism for early stopping if convergence is achieved.

Table 6 displays the pseudo code of early stopping criteria if convergence is achieved.

Table 6. Pseudo code for convergence threshold

```
Function has_converged(best_fitness_history, stagnation_limit):
    # Check if the length of best_fitness_history is less than the stagnation_limit
    If length(best_fitness_history) < stagnation_limit Then
        Return False # Convergence not reached yet
    End If

    # Check if the last 'stagnation_limit' elements in best_fitness_history are all the same
    If all(best_fitness_history[-stagnation_limit:] == best_fitness_history[-1]) Then
        Return True # Convergence has been reached
    Else
        Return False # Convergence not reached yet
    End If
End Function
```

- The optimal solution to the problem is generally regarded as being the best solution discovered during the optimization process.

Directly expressing the distribution as a simple analytical formula is often infeasible due to the complexity of real-world optimization landscapes. Instead, the invariant distribution of solutions is determined empirically through the optimization process itself. The algorithm explores the solution space by iteratively generating and evaluating candidate solutions, adapting its parameters, and gradually converging towards regions of interest in the search space.

Fig. 2 displays a visual representation of where the algorithm found solutions in the search space defined by the x and y axes. Areas with a higher frequency of solutions (elevated regions in the 3D plot) indicate that the algorithm frequently finds solutions in those areas. This suggests that those regions contain multiple local optima. The peak that stands out from the rest suggests the location of the global optimum. Multiple peaks in the plot indicate the presence of local optima, and the algorithm has found solutions in these local optima. Table 7 displays the pseudocode of the algorithm for non-convex sinusoidal function.

Table 7. Pseudo code Non-Convex Objective Function (Sinusoidal Function)

```
Function sinusoidal_function(x):
    Return sin(x[0]) + cos(x[1])
End Function
```

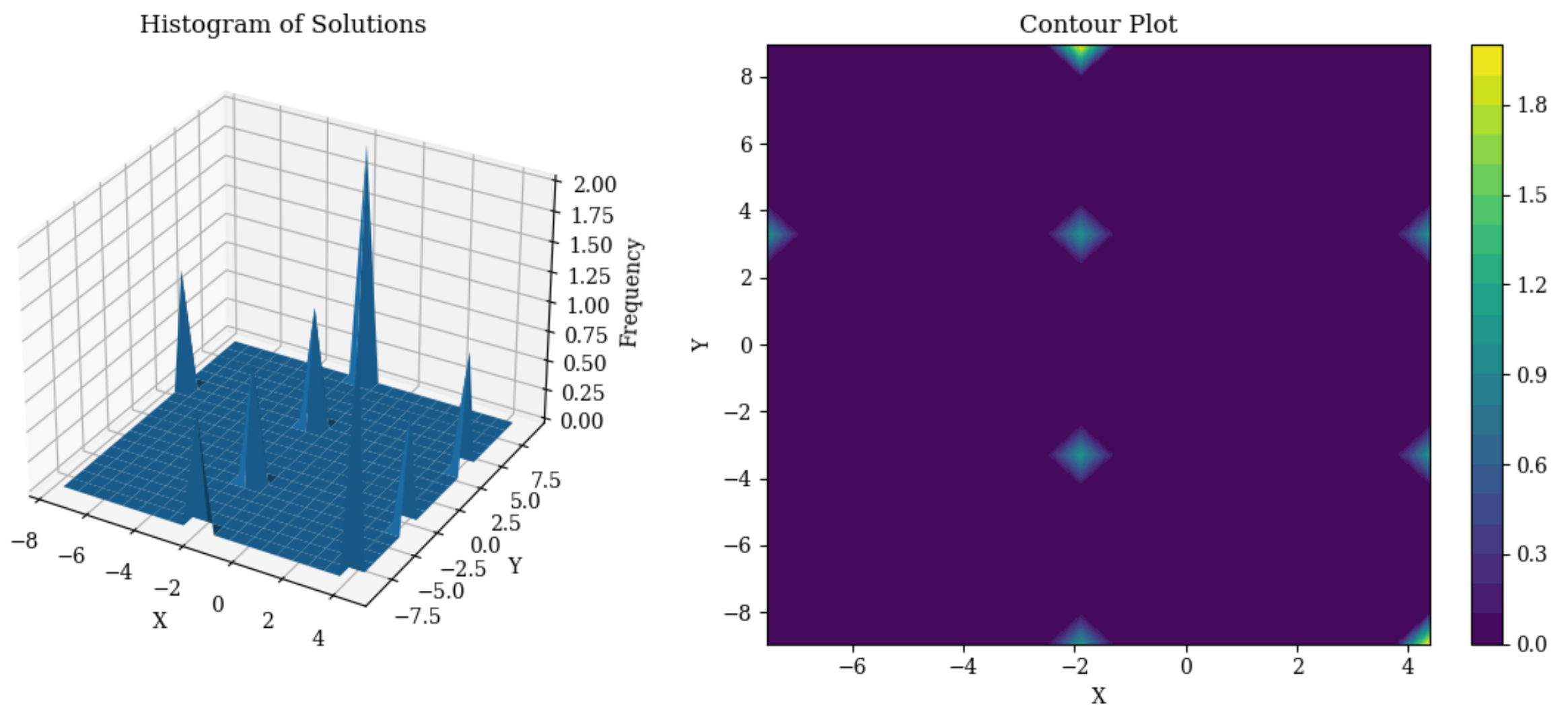

Fig. 2. Solution Distribution and Density

ADEDS algorithm is configured to run with a population size of 50, a maximum of 100 generations, and it's restricting the search space to the range -10 to 10 for both x and y dimensions. The algorithm tries to find the optimal solution within these constraints.

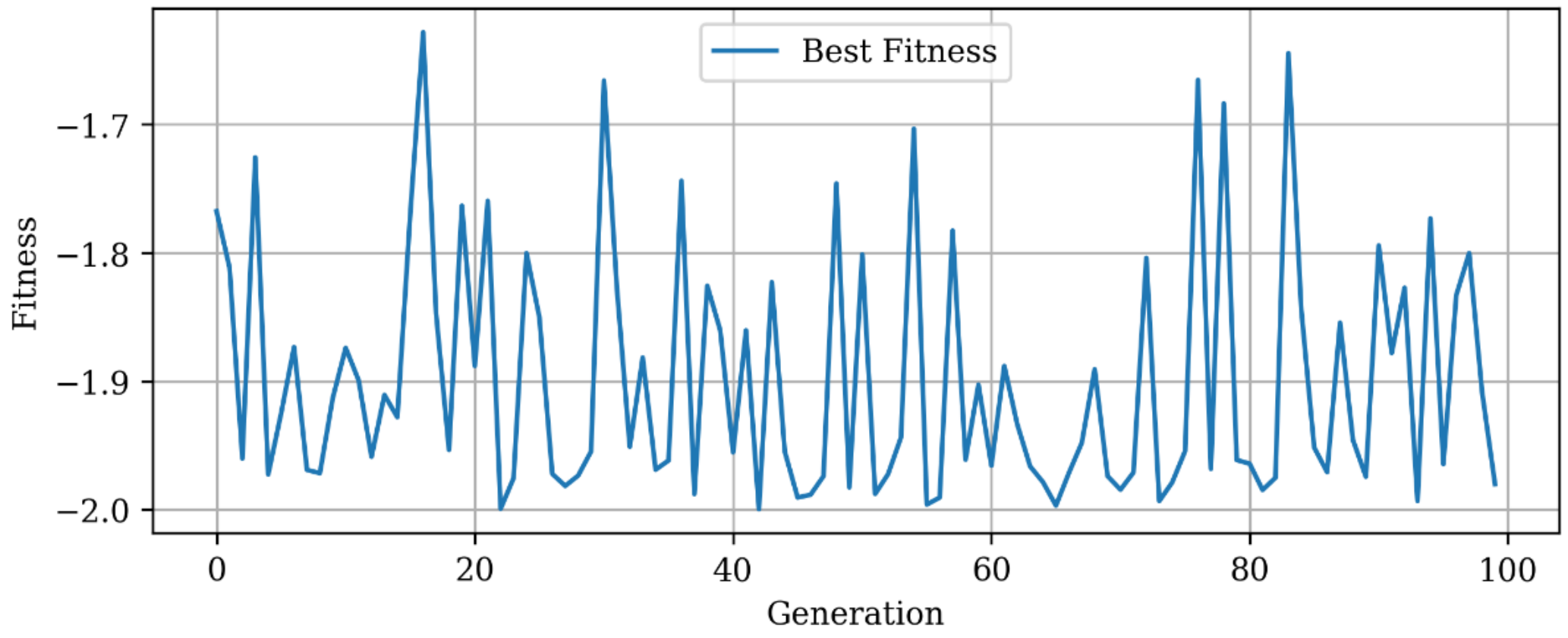

Fig. 3. ADEDS Convergence plot: Sinusoidal objective function

Fig 3 displays the convergence where; we see that the fitness value decreases as the number of generations increases. This indicates that the optimization algorithm is converging towards a better solution over time. The decreasing trend suggests that the algorithm is effective in finding improved solutions at each generation.

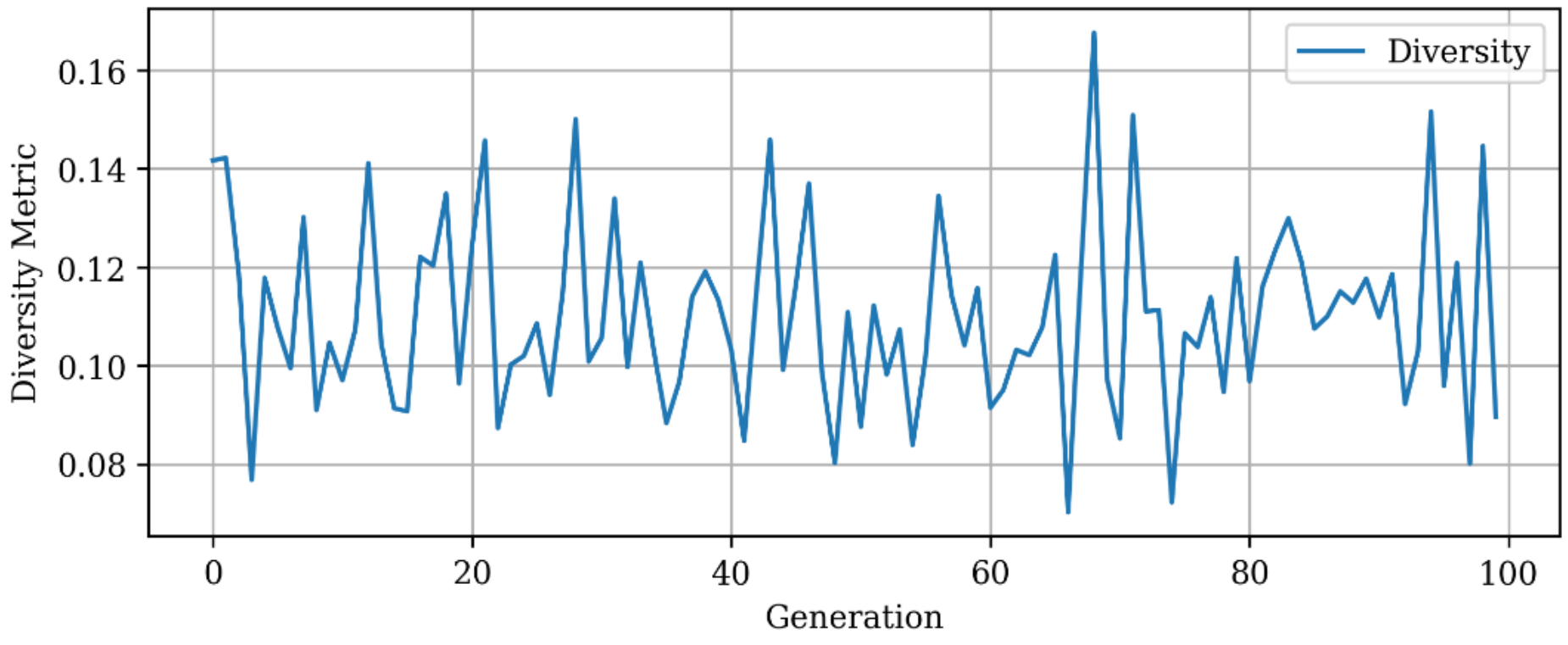

Fig. 4. Diversity: ADEDS with Sinusoidal objective function

Fig. 4 displays the diversity and how the diversity metric evolves over generations in optimization. It starts with high diversity, indicating a wide range of solutions, decreasing as generations progress. Eventually, it reaches a stable level, indicating a limited set of solutions.

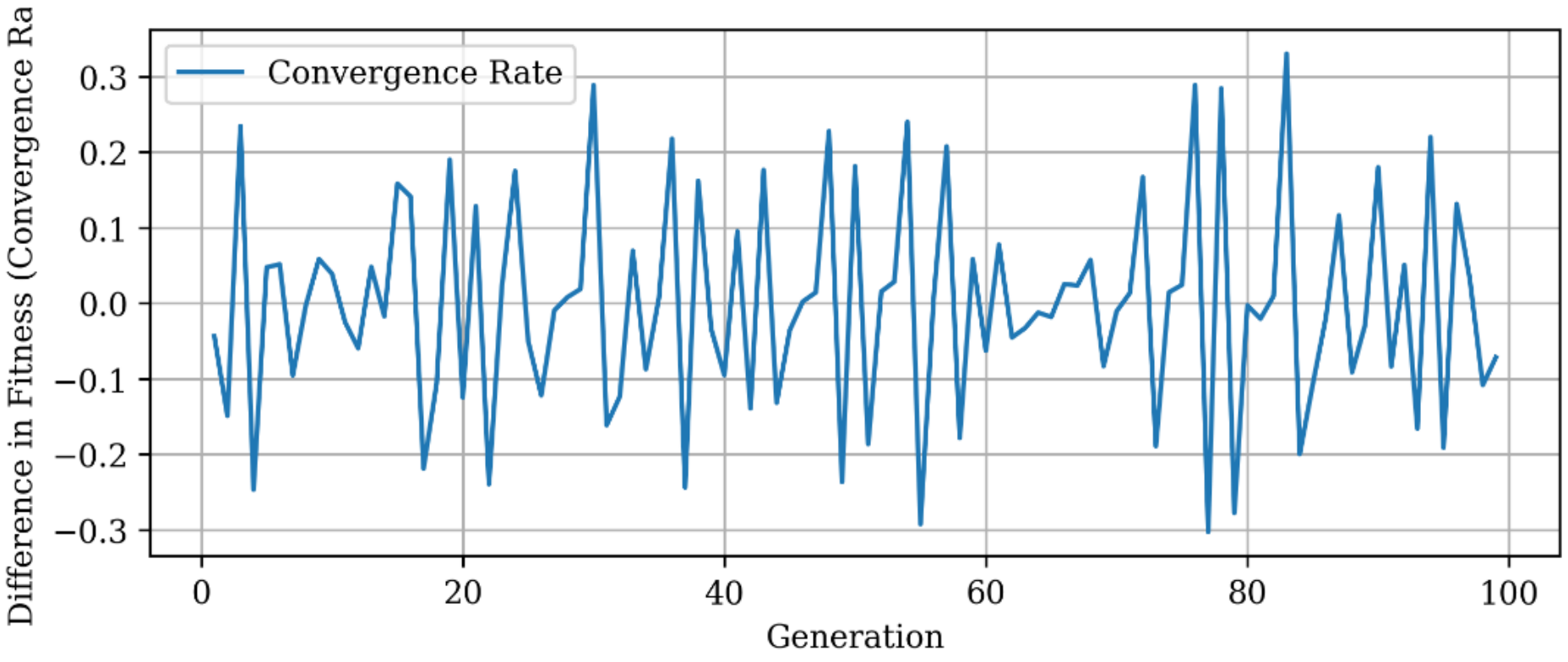

Fig. 5. Convergence Rate

Fig. 5 displays the convergence rate, where we see the improvement in the algorithm, with negative values indicating decreasing fitness values and larger values indicating faster convergence. As the process progresses, significant improvements become less frequent.

These three plots collectively illustrate the optimization process's dynamics and efficiency. The convergence plot demonstrates that the algorithm is consistently finding better solutions over time. The Diversity Plot shows how the population's variety decreases as the algorithm converges to a narrower set of solutions. The convergence rate plot quantifies the speed of improvement, with rapid convergence at the early stages of optimization. Table 8 displays the optimal solution and convergence rate for the 10-D problem.

Table 8. Optimal solution and convergence rate

| **Best solution (10 dimensions)** | **Best fitness** | **Final convergence rate** |
|---|---|---|
| [36.14, -204.24, 104.18, -39.79, -18.74, 65.49, -35.66, -15.51, 61.75, -96.77] | -1.9992 | -0.0715 |

The algorithm finds a low fitness value (close to -2.0) for the best solution, indicating a near-optimal or optimal solution. The final convergence rate was close to zero, indicating a stabilized search. The best fitness value is close to the global optimum, indicating successful performance.

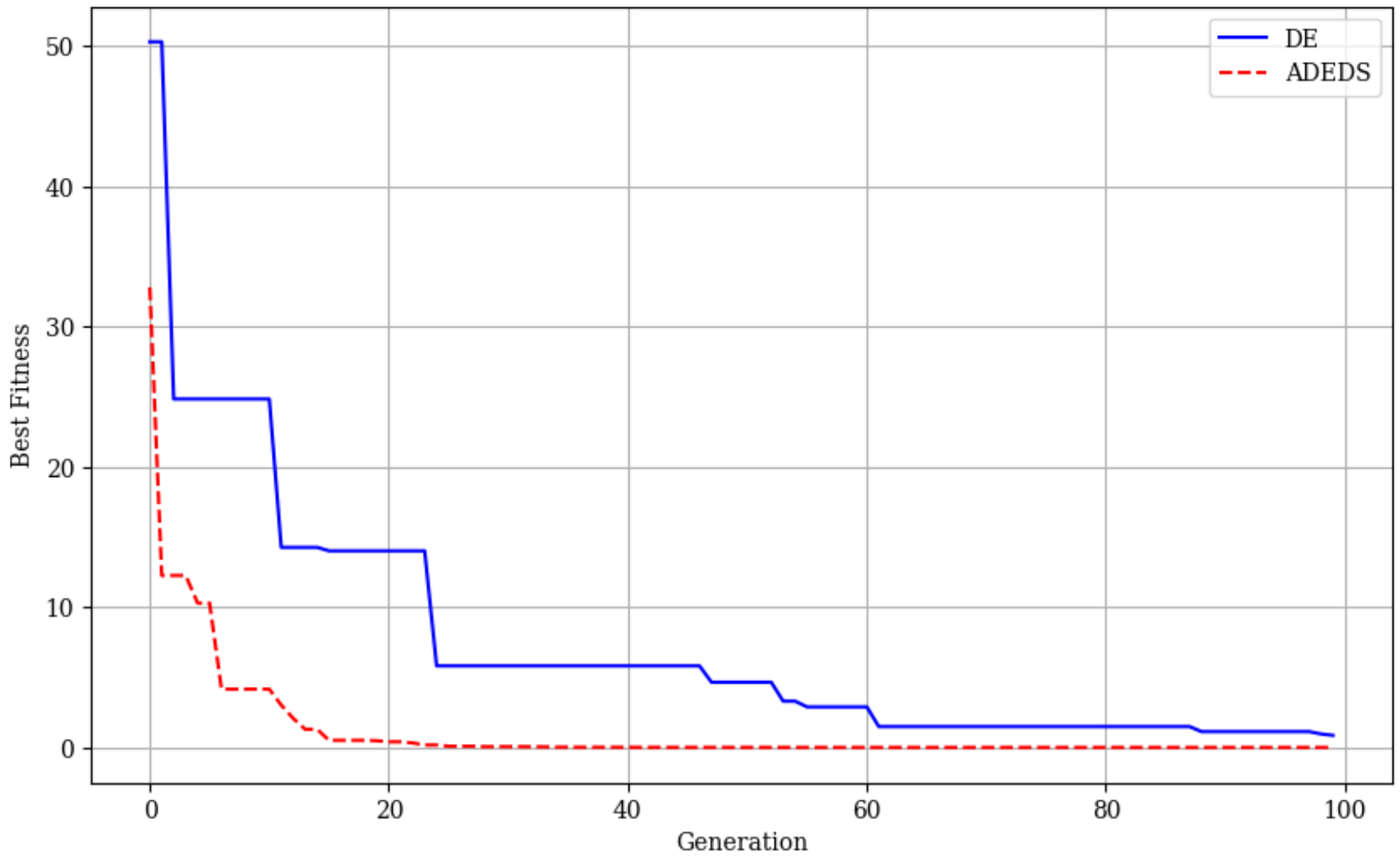

Fig. 6. Comparison with traditional DE (sphere function)

The algorithm that converges to a lower fitness value faster is considered more efficient. Fig. 6 displays that the ADEDS line is lower than the DE line at a particular generation, which means that the ADEDS algorithm found a better solution (lower fitness) at that generation when applied to the convex objective function. However, ADEDS's aim is to find the global minimum efficiently in more complex and non-convex landscapes. Fig. 7 displays the convergence of the sinusoidal objective function.

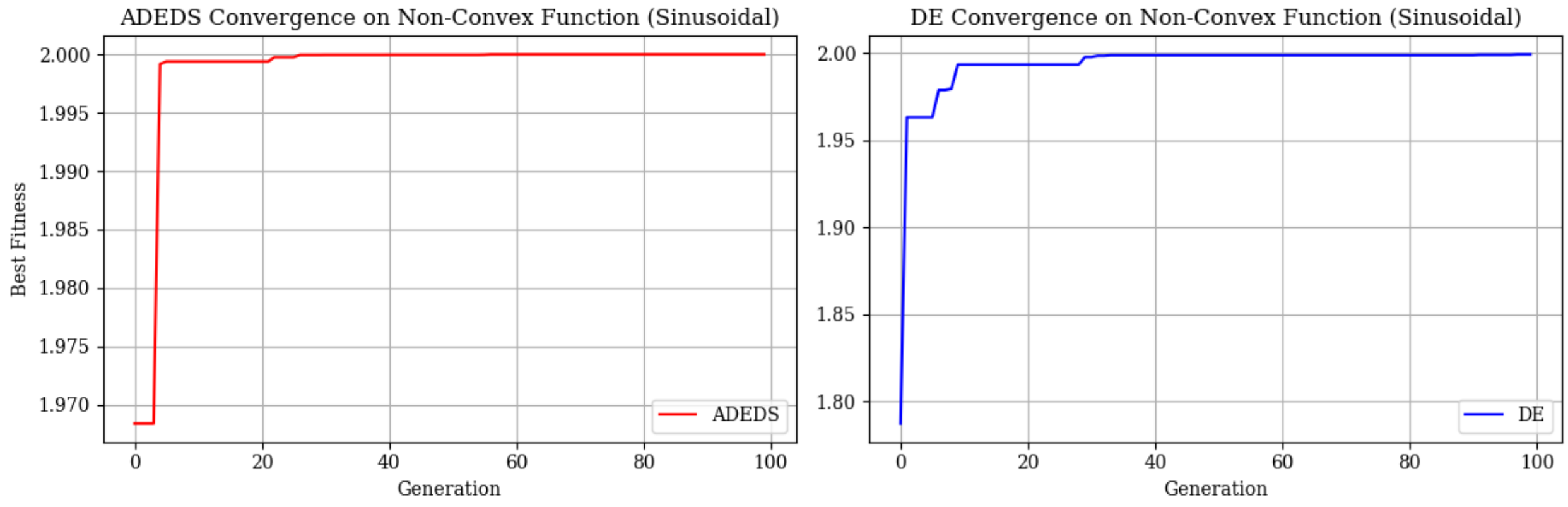

Fig. 7. Convergence behavior with sinusoidal objective function

Table 9 displays a comparative study between the two.

Table 9. Comparative analysis

| DE | ADEDS |
|---|---|
| Typically uses fixed values for its parameters (F and CR) throughout the optimization process. | Dynamically adjusts its key parameters (F- scaling factor and CR- crossover rate) during optimization based on the current state of the search. It adapts to changing conditions in the landscape. |
| Uses a fixed strategy for mutation and crossover, which may not adapt to different stages of optimization. | Employs an adaptive strategy for parameter adjustment. It monitors fitness values to adapt F and CR. This allows it to explore and exploit the search space more effectively. |
| Selects individuals based on fitness, favoring better fitness values. | Selects individuals based on fitness but may prioritize maintaining diversity in the population to aid exploration. |
| Does not adapt its parameter values, which can lead to premature convergence or slow convergence in challenging landscapes. | Continuously adapts its parameters and strategies, allowing it to navigate complex landscapes more effectively and avoid getting stuck in local optima. |
| Terminates based on predefined criteria such as the number of generations or convergence. | Terminates based on predefined criteria but has a better chance of reaching the global optimum due to its adaptability. |
| Struggle to escape local optima in complex landscapes, as it lacks adaptability in parameter settings. | Excels in complex landscapes with multiple local optima. Its adaptability allows it to dynamically adjust its strategy and parameters to explore and exploit the search space effectively, increasing the chances of finding the global optimum. |

While traditional DE can be effective for optimizing convex functions, its limitations become apparent when dealing with non-convex problems, where it struggles to escape local optima. Fig. 9 displays the DE optimization for both convenx and non-convex optimization. In the convex problem, the objective function is $f_{convex}(x) = x_0^2 + x_0^2$. The objective function is smooth and has a single optimal solution at the center, which DE successfully finds. In the non-convex problem, the Sinusoidal objective function is $f_{nonconvex}(x) = \sin(x_0) + \sin(x_1)$ where DE struggles to find the global minima due to the presence of multiple local optima. This function presents a landscape where the algorithm is searching for the minimum value. However, there are many points in the search space where the gradient (derivative) of the function is zero, indicating a potential local minimum.

$$\nabla f_{nonconvex}(x_0^* + x_1^*) = \left(\frac{\partial f_{nonconvex}}{\partial x_0}(x_0^* + x_1^*),\ \frac{\partial f_{nonconvex}}{\partial x_1}(x_0^* + x)\right) = (0,0)$$

while DE converged to a solution $(x_0^* + x_1^*)$, it may not be the best possible solution due to the presence of potentially better global optimum elsewhere in the landscape.

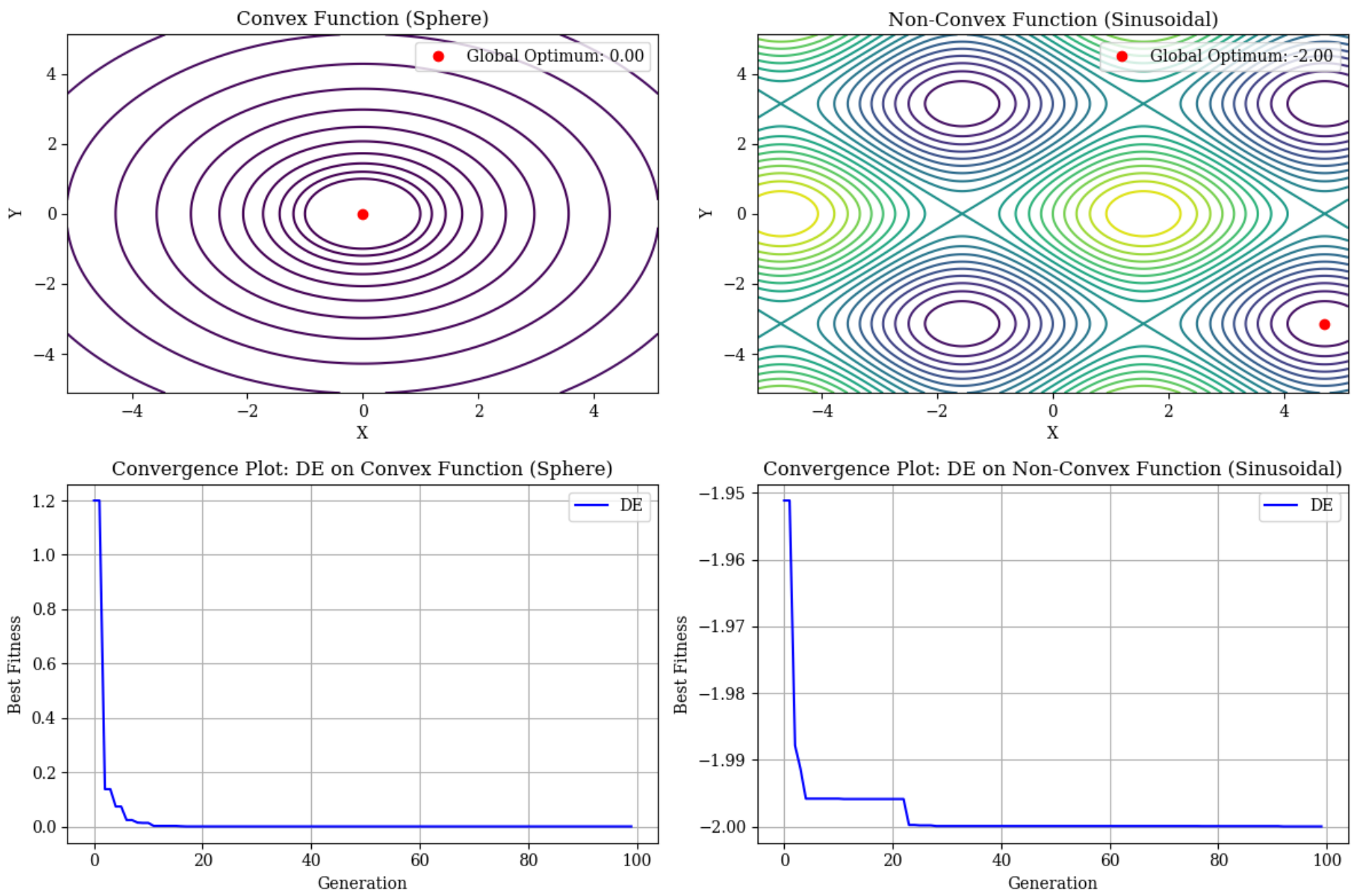

Fig. 8 DE Optimization Results for Convex and Non-Convex Problems

$$f(x_0^* + x_1^*) \leq \min_{global} f(x_0 + x_1)$$

Where $\min_{global} f(x_0 + x_1)$ represents the true global minima. When we apply the ADEDS algorithm to this problem, it efficiently explores the search space, adaptively adjusting its parameters like mutation scale factor (F) and crossover rate (CR) to guide the search.

To assess the algorithm's performance and applicability for various optimization challenges, we tested it against a variety of benchmark functions. Appendix 1 displays the success rate for various test functions with the numbers of iterations.

## 3. Benchmark functions

Benchmark functions are crucial for evaluating and developing optimization methods, providing a controlled environment for assessing algorithm performance. They cover various optimization challenges, such as convex and non-convex landscapes, single and multiple optima, and multimodality levels. Test functions are classified based on surface shape, such as bowl-shaped, plate-shaped, and multiple local minima, representing increasing difficulty in optimizing test functions.

### 3.1 Many local optima

The gradient-based optimization technique struggles to find the minima of the functions with multiple local optima. This can be solved using heuristic search-based optimization methods. These algorithms begin with several starting points and end with solutions.

#### 3.1.1 Ackley function

Ackley function (Ackley, 1987) is a multimodal and non-convex function with multiple local minima and a global minimum at $f(x) = 0$, which occurs when $x$ is the zero vector (Back, 1996).

$$f_{ackley}(x) = -a\ exp\left(-b\sqrt{\frac{1}{n}\sum_{i=1}^{n} x_i^2}\right) - exp\left(\frac{1}{n}\sum_{i=1}^{n} cos(2\pi x_i)\right) + a + exp(1)$$

Where, n is the dimension of vector x, $\sum_{i=1}^{n} x_i^2$ represents the sum of the squares of all components of the vector x, $\sum_{i=1}^{n} cos(2\pi x_i)$ represents the sum of the cosine values of each component of x after scaling by $2\pi$. The recommended variable values are a = 20, b = 0.2 and c = 2π. Here, $a\ exp\left(-b\sqrt{\frac{1}{n}\sum_{i=1}^{n} x_i^2}\right)$ depends on the Euclidean norm (root mean square) of x. It decreases exponentially as the norm increases. $exp\left(\frac{1}{n}\sum_{i=1}^{n} cos(2\pi x_i)\right)$ is the sum of the cosine values of each component. It oscillates between -1 and 1, contributing to the function's oscillatory behavior. The constants and exp (1) are added to shift the function to a minimum value of 0 at the global minimum. Here global minima $f_{ackley}(0,0) = 0.00$.

Table 10. Pseudo code Ackley function

```
function Ackley (x):
   dimension = length (x)
   sum_sq = 0
   sum_cos = 0

   for each element in x:
      sum_sq = sum_sq + element ^ 2
      sum_cos = sum_cos + cos (2 * π * element)

   term1 = -20 * exp (-0.2 * sqrt (sum_sq / dimension))
   term2 = -exp (sum_cos / dimension)

   result = term1 + term2 + 20 + exp (1)

   return result
```

### 3.1.2 Bukin function N.6

The region around Bukin functions' minimal points resembles a fractal (with subtle seesaw edges). This characteristic makes them extremely challenging to optimize using any global (or local) optimization technique.

$$f_{bukin}(x) = 100\sqrt{x_2 - 0.01x_1^2} + 0.01(x_1 + 10)$$

Table 11. Pseudo code Bukin Function

```
function Bukin(x):
   x, y = x
   term1 = 100 * sqrt (abs (y - 0.01 * x^2))
   term2 = 0.01 * abs (x + 10)
   result = term1 + term2
   return result
```

The function validated on the rectangle $x_1\ \epsilon\ [-15,-5],\ x_2\ \epsilon\ [-3,-3]$. Here global minima $f_{bukin}(x^*) = 0.00$.

### 3.1.3 Rastrigin function

The Rastrigin function is a classic example of a multi-modal optimization problem (Rastrigin, 1974).

$$f_{rastrigin}(x) = 10n + \sum_{i=1}^{n} [x_i^2 - 10Cos(2\pi x_i)]$$

Table 12. Pseudo code for Rastrigin function

```
function Rastrigin(x):
   A = 10
   sum_of_squares = 0
   for xi in x:
      sum_of_squares += xi ^ 2 - A * cos (2 * pi * xi)
   result = A * length(x) + sum_of_square
  return result
```

The function validated on the hypercube $x_i \in [-5.12, 5.12], for\ all\ i = 1, \ldots, n.$ Here global minima$f_{rastrigin}(0, \ldots, 0) = 0.00.$

### 3.1.4 Cross-in-tray function

Cross-in-Tray function has multiple global minima (Jamil & Yang, 2013).

$$f_{crossintray}(x) = -0.0001\left(sin(x_1)sin(x_2)exp\left(100 - \frac{\sqrt{x_1^2 + x_2^2}}{\pi}\right) + 1\right)^{0.1}$$

Table 13. Pseudo code for Cross-in-Tray function

```
function CrossInTray(x):
   x, y = x
   a = abs (100 - sqrt (x^2 + y^2) / pi)
   b = abs(sin(x) * sin(y) * exp(a)) + 1
   result = - 0.0001 * (abs (sin (x) * sin (y) * exp (a)) + 1) ^ 0.1
   return result
```

The function validated on:

$$min = \begin{cases} f_{crossintray}(1.35, -1.35) = 2.06262 \\ f_{crossintray}(1.35, 1.35) = 2.06262 \\ f_{crossintray}(-1.35, 1.35) = 2.06262 \\ f_{crossintray}(-1.35, -1.35) = 2.06262 \end{cases}, \text{with search domain } -10 \leq x, y \leq 10$$

### 3.1.5 Levy function

$$f_{levy}(x) = sin^2(\pi w_1) + \sum_{i}^{n-1} (w_i - 1)^2 [1 + 10sin^2(\pi w_i + 1)] + (w_n - 1)^2[1 + sin^2(2\pi w_n)]$$

Where, $\mathrm{w_i} = 1 + \frac{x_{\mathrm{i}} - 1}{4}$, for all $\mathrm{i} = 1, \ldots, \mathrm{n}$.

Table 14. Pseudo code for Levy function

```
function Levy(x):
    x, y = x
    term1 = sin (3 * pi * x) ^2
    term2 = (x - 1) ^2 * (1 + sin (3 * pi * y) ^2)
    term3 = (y - 1) ^2 * (1 + sin (2 * pi * y) ^2)
    result = term1 + term2 + term3
    return result
```

The function validated on the hypercube $x_\mathrm{i} \in [-10, 10],\ for\ all\ i = 1, \ldots, n$. Here global minima $f_{levy}(1, 1) = 0.00$.

### 3.1.6 Egg-holder function

The topography of the non-convex Egg-Holder function is misleading, and it is an incredibly difficult function to optimize (Whitley et al., 1996).

$$f_{eggholder}(x) = -(x_2 + 47)sin\left(\sqrt{x_2 + \frac{x_1}{2} + 47}\right) - x_i sin\left(\sqrt{x_1 - (x_2 + 47)}\right)$$

Table 15. Pseudo code for Egg-Holder function

```
function EggHolder(x):
    x, y = x
    a = sqrt (fabs (y + x/2 + 47))
    b = sqrt (fabs (x - (y + 47)))
    result = - (y + 47) * sin (a) - x * sin (b)
    return result
```

This function validated on the square $x_i \in [-512, 404.2319], for\ all\ i = 1, 2$. Here global minima $f_{eggholder}(x^*) = -959.6407$.

### 3.1.7 Schaffer function N. 2

$$f_{schaffer}(x) = 0.5 + \frac{sin^2(x_{1-}^2 x_2^2) - 0.5}{[1 + 0.001(x_{1-}^2 x_2^2)]^2}$$

Table 16. Pseudo code Schaffer function

```
function Schaffer(x):
    x, y = x
    numerator = sin (x^2 - y^2) - 0.5
    denominator = (1 + 0.001 * (x^2 + y^2)) ^2
    result = 0.5 + (numerator / denominator)
    return result
```

This function validated on the square $x_i \in [-100, 100], for\ all\ i = 1, 2$. Here global minima $f_{schaffer}(x^*) = 0.00$.

### 3.1.8 Schwefel function

The Schwefel function is complex, with many local minima (Schwefel, 1981)**.**

$$f_{schwefel}(x, y) = 418.9829 \ * \ 2 - (x * sin(\sqrt{|x|})) + -(y * sin(\sqrt{|y|}))$$

This function validated on the hypercube $x_i \in [-500, 500], for \ all \ i = 1, \dots, n$. Here global minima $f_{schwefel}(x^*) = 0.00$.

Table 17. Pseudo code Schwefel function

```
function Schwefel(x):
    x, y = x [0], x [1]
    result = 418.9829 * 2 - (x * sin (sqrt (abs (x)))) - (y * sin (sqrt (abs (y))))
    return result
```

### 3.1.9 Shubert function

This function has several local minima and many global minima.

$$f_{shubert}(x) = \left(\sum_{i=1}^{5} i \ cos((i+1)x_1 + i)\right)\left(\sum_{i=1}^{5} i \ cos((i+1)x_2 + i)\right)$$

Table 18. Pseudo function Shubert function

```
function Shubert(x):
    x, y = x
    m = 5
    result = 1
    for i in range (1, m + 1):
        temp_sum = sum ((2 * i) * cos ((2 * i) * (x + y)))
        result *= temp_sum
    return result
```

This is validated on the square $x_i \in [-10, 10], for \ all \ i = 1, 2$. Here global minima $f_{shubert}(x^*) = -186.7309$.

### 3.1.10 Drop-Wave function

This is multimodal and highly complex function.

$$f_{dropwave}(x) = -\frac{1 + cos\left(12\sqrt{x_1^2 + x_2^2}\right)}{0.5(x_1^2 + x_2^2) + 2}$$

Table 19. Pseudo code for Drop-Wave function

```
function DropWave(x):
    x, y = x
    numerator = - (1 + cos (12 * sqrt (x^2 + y^2)))
    denominator = 0.5 * (x^2 + y^2) + 2
    result = numerator / denominator
    return result
```

It is validated on the square $x_i \in [-5.12, 5.12], for \ all \ i = 1, 2$. Here global minima $f_{dropwave}(x^*) = -1$.

### 3.1.11 Himmelblau's function

It is a multi-modal function (Himmelblau, 2018).

$$f_{himmelnlau}(x, y) = (x^2 + y - 11)^2 + (x + y^2 - 7)^2$$

Table 20. Pseudo code Himmelblau

```
function Himmelblau(x):
    x, y = x
    term1 = (x^2 + y - 11) ^2
    term2 = (x + y^2 - 7) ^2
    result = term1 + term2
    return result
```

$$min = \begin{cases} f_{himmelnlau}(3.00{,}2.00) = 0 \\ f_{himmelnlau}(-2.80, 3.13) = 0 \\ f_{himmelnlau}(-3.78, -3{,}28) = 0 \\ f_{himmelnlau}(3.58, -1.84) = 0 \end{cases}, \text{ with search domain } -5 \le x, y \le 5$$

## 3.2 ADEDS with many local optima

Many local optima in optimization problems make it difficult for algorithms to locate the global optimum, which is the best feasible solution to the problem. Table 21 reports ADEDS with the many local optima. All the tests are performed using two dimensions, 50 population sizes, and 10 iterations. We have run the optimization algorithms 10 times for each function (num_runs), which determines that independent optimization runs will be conducted to compare the performance of DE and ADEDS. Each run initializes the algorithm independently and runs it for 50 iterations (max_generations). It determines how many iterations each algorithm will perform before stopping.

Significant differences in mean fitness values and low p-values suggest that ADEDS outperform DE on several benchmark functions except Shubert and Himmelblau functions. In all the test functions, ADEDS achieved the global minima with just 10 runs, which is promising. Grid-based parameter setting can be employed here to optimize the performance of algorithms for specific problems.

Table 21. ADEDS with many local optima benchmark test report

| Sl no | Benchmark function | | Mean fitness | Std dev | t-stats | p-value |
|---|---|---|---|---|---|---|
| 1 | Rastrigin | DE | 8.429 | 3.733 | -6.773 | 0.000*** |
| | | ADEDS | 0.000 | 0.000 | | |
| 2 | Ackley | DE | 11.460 | 3.913 | -8.784 | 0.000*** |
| | | ADEDS | 0.000 | 0.000 | | |
| 3 | Cross in tray | DE | -1.935 | 0.065 | -3.905 | 0.001*** |
| | | ADEDS | -2.062 | 0.000 | | |
| 4 | Egg holder | DE | -1446.82 | 372.62 | 3.922 | 0.000*** |
| | | ADEDS | -959.640 | 0.000 | | |
| 5 | Drop-wave | DE | -0.974 | 0.031 | -2.449 | 0.024** |
| | | ADEDS | -1.000 | 0.000 | | |
| 6 | Levy | DE | 4.379 | 3.365 | -3.904 | 0.001*** |
| | | ADEDS | 0.000 | 0.000 | | |
| 7 | Schwefel | DE | -0.0732 | 0.223 | 2.930 | $0.008^{***}$ |
| | | ADEDS | -0.995 | 0.000 | | |
| 8 | Schaffer | DE | -0.073 | 0.223 | -12.380 | $0.000^{***}$ |
| | | ADEDS | -0.995 | 0.000 | | |
| 9 | Bukin | DE | 25.450 | 5.211 | -14.651 | $0.000^{***}$ |
| | | ADEDS | 0.000 | 0.000 | | |
| 10 | Shubert | DE | -3292.422 | 624.517 | -3.044 | $0.010^{***}$ |
| | | ADEDS | -3840.000 | 0.000 | | |
| 11 | Himmelblau | DE | 0.000 | 0.000 | 2.627 | $0.017^{***}$ |
| | | ADEDS | 0.000 | 0.000 | | |

### 3.3 Plate-Shaped

Shape optimization is a well-established part of the calculus of variations. It is an extension of optimal control theory, with the minimizing parameter being simply the domain in which the problem is described. There are frequently several equal or nearly similar solutions for plate-shaped functions. Since the objective function values of these solutions are so similar, it is challenging for optimization algorithms to discriminate between them and choose the optimal one

#### 3.3.1 Booth Function

$$f_{booth}(x) = (x_1 + 2x_2 - 7)^2 + (2x_1 + x_2 - 5)^2$$

Table 22. Pseudo code Booth function

```
function Booth(x):
    x, y = x
    term1 = (x + 2*y - 7) ^2
    term2 = (2*x + y - 5) ^2
    result = term1 + term2
    return result
```

This is validated on the square $x_i \in [-10, 10], for\ all\ i = 1, 2.$ Here global minima $f_{booth}(x^*) = 0.00.$

#### 3.3.2 Matyas Function

This test function has no local minima.

$$f_{matyas}(x) = 0.26(x_1^2 + x_2^2) - 0.48x_1x_2$$

Table 23. Pseudo code Matyas function

```
function Matyas (x):
    x, y = x
    term1 = 0.26 * (x^2 + y^2)
    term2 = - 0.48 * x * y
    result = term1 + term2
    return result
```

It is validated on the square $x_i \in [-10, 10], for\ all\ i = 1, 2.$ Here global minima $f_{matyas}(0, 0) = 0.00.$

### 3.3.3 McCormick Function

$$f_{mccormick}(x) = sin(x_1 + x_2) + (x_1 + x_2)^2 - 1.5x_1 + 2.5x_2 + 1$$

Table 24. Pseudo code McCormick function

```
function McCormick(x):
    x, y = x
    term1 = sin (x + y)
    term2 = (x - y) ^2
    term3 = -1.5 * x
    term4 = 2.5 * y
    result = term1 + term2 + term3 + term4 + 1
    return result
```

This is validated on the rectangle $x_1 \in [-1.5, 4], x_2 \in [-3, 4]$. Here global minima $f_{mccormick}(-0.54719, -1.54719) = -1.9133$.

### 3.2 ADEDS with plate shaped.

Table 23 presents the performance report, which clearly shows the superiority of ADEDS over DE in all categories. Here too, all the tests are performed using the same configurations. We see that ADEDS achieves the global optima in all three tests with just 10 runs.

Table 25. ADEDS with plate shaped benchmark test report

| **Sl no** | **Benchmark function** | | **Mean fitness** | **Std dev** | **t-stats** | **p-value** |
|---|---|---|---|---|---|---|
| 1 | McCormick | DE | -1.913 | 2.230 | 4.113 | $0.000^{***}$ |
| | | ADEDS | -4.972 | 0.000 | | |
| 2 | Matyas | DE | 0.585 | 0.757 | -2.318 | $0.004^{***}$ |
| | | ADEDS | 0.000 | 0.000 | | |
| 3 | Booth | DE | 10.417 | 0.000 | -6.309 | $0.000^{***}$ |
| | | ADEDS | 0.000 | 0.000 | | |

### 3.3 Valley-Shaped

Long, narrow valleys with several local optima are common in valley-shaped functions. Optimization algorithms may struggle to avoid these valleys, resulting in convergence to suboptimal solutions.

### 3.3.1 Three-Hump Camel Function

$$f_{threehumpcamel}(x) = 2x_1^2 - 1.05x_1^4 + \frac{x_1^6}{6} + x_1x_2 + x_2^2$$

Table 26. Pseudo code for Three-Hump Camel function

```
function ThreeHumpCamel(x):
    x, y = x
    term1 = 2 * x^2
    term2 = -1.05 * x^4
    term3 = (x^6) / 6
    term4 = x * y
    term5 = y^2
    result = term1 + term2 + term3 + term4 + term5
    return result
```

This is validated on the square $x_i \in [-5,5], for\ all\ i = 1,2$. Here global minima $f_{threehumpcamel}(x^*) = 0$.

### 3.3.2 Six-Hump Camel Function

$$f_{sixhumpcamel}(x) = \left(4 - 2.1x_1^2 + \frac{x_1^4}{3}\right)x_1^2 + x_1x_2 + (-4 + 4x_2^2)x_2^2$$

Table 27. Pseudo code (Six-hump Camel)

```
function SixHumpCamel(x):
    x, y = x
    term1 = 4 - 2.1 * x^2 + (x^4) / 3
    term2 = x^2
    term3 = x * y
    term4 = -4 + 4 * y^2
    term5 = y^2
    result = term1 * term2 + term3 + term4 * term5
    return result
```

This is validated on the rectangle $x_1 \in [-3,3]$, $x_2 \in [-2,2]$. Here global minima $f_{sixhumpcamel}(x^*) = 1.0316$.

### 3.3.3 Rosenbrock Function

The global minimum of the unimodal function is in a small, parabolic valley (Rosenbrock, 1960). Convergence to the minimum is challenging, even though this valley is simple to locate (Picheny et al., 2012).

$$f_{rosenbrock}(x) = \sum_{i}^{n-1}[100(x_{i+1} - x_i^2)^2 + (x_i - 1)^2]$$

The is validated on the hypercube $x_i \in [-5,10], for\ all\ i = 1,\dots,d$, although it may be restricted to the hypercube xi ∈ [-2.048, 2.048], for all i = 1, …, n. Here global minima $f_{rosenbrock}(x^*) = 0$.

Table 28. Pseudo code Rosenbrock function

```
function Rosenbrock(x):
    sum = 0
    for i in range (length (x) – 1):
        term1 = 100.0 * (x [i + 1] – x [i] ** 2) ** 2
        term2 = (1 – x [i]) ** 2
        sum += term1 + term2
    return sum
```

$$min = \begin{cases} n = 2 \rightarrow f_{rosenbrock}(1{,}1) = 0 \\ n = 3 \rightarrow f_{rosenbrock}(1{,}1{,}1) = 0 \\ n > 3 \rightarrow f_{rosenbrock}(1,\ldots,1) = 0 \end{cases}, \text{with search domain } -\infty \leq x_i \leq \infty, 1 \leq i \leq n$$

### 3.3.4 Dixon-Price function

$$f_{dixonprice}(x) = (x_i - 1)^2 + \sum_{i=2}^{n} [i(2x_i^2 - x_{i-1})^2]$$

Table 29. Pseudo code Dixon-Price function

```
function Dixon_Price(x):
    x, y = x
    term1 = (x - 1) ** 2
    term2 = (y - 1) ** 2
    term3 = 100 * (x ** 2 - y) ** 2
    return term1 + term2 + term3
```

The is validated on the hypercube $x_i \in [-10, 10], for\ all\ i = 1, \ldots, n$. Here global minima $f_{dixonprice}(x^*) = 0$.

Table 27 displays the report with clear superiority of ADEDS over DE in all categories. Rosenbrock function is unimodal. Its minimum is tucked away in a valley with a flat bottom and a banana-shaped shape. Here, the algorithm required more than 100 runs to succeed.

Table 30. ADEDS with valley-shaped benchmark test report

| Sl no | Benchmark function | | Mean fitness | Std dev | t-stats | p-value |
|---|---|---|---|---|---|---|
| 1 | Rosenbrock | DE | 50.138 | 53.356 | -2.819 | 0.011*** |
| | | ADEDS | 0.000 | 0.000 | | |
| 2 | Three-hump camel | DE | 0.784 | 0.481 | -4.883 | 0.000*** |
| | | ADEDS | 0.000 | 0.001 | | |
| 3 | Six-hump camel | DE | -0.759 | 0.167 | -4.865 | 0.000 |
| | | ADEDS | -1.031 | 0.000 | | |
| 3 | Dixon price | DE | 65.246 | 50.986 | -3.839 | 0.001*** |
| | | ADEDS | 0.000 | 0.000 | | |

## 3.4 Other

### 3.4.1 Beale Function

This is multimodal, with sharp peaks at the corners of the input domain.

$$f_{beale}(x) = (1.5 - x_1 + x_1x_2)^2 + (2.25 - x_1 + x_1x_2^2)^2 + (2.625 - x_1 + x_1x_2^3)^2$$

Table 31. Pseudo code (Beale shaped)

```
function Beale(x):
    term1 = (1.5 - x [0] + x [0] * x [1]) ^2
    term2 = (2.25 - x [0] + x [0] * x [1] ^2) ^2
    term3 = (2.625 - x [0] + x [0] * x [1] ^3) ^2
    return term1 + term2 + term3
```

It is validated on the square $x_i \in [-4.5, 4.5], for\ all\ i = 1, 2$. Here global minima $f_{beale}(3, 0) = 0$.

### 3.4.2 Goldstein-Price Function

This function has several local minima.

$$f_{goldsteinprice}(x) = [1 + (x_1 + x_2 + 1)^2 + (19 - 14x_1 + 3x_1^2 - 14x_2 + 6x_1x_2 + 3x_2^2)] * [(30 + (2x_1 + 3x_2)^2(18 - 32x_1 + 12x_1^2 + 48x_2 - 36x_1x_2 + 27x_2^2)]$$

It is validated on the square $x_i \in [-2, 2], for\ all\ i = 1, 2$. Here global minima $f_{goldsteinprice}(0, -1) = 3$.

Table 32. Pseudo code Goldstein-Price function

```
function Goldstein-Price(x):
    term1 = (1 + (x [0] + x [1] + 1) ^2 * (19 - 14 * x [0] + 3 * x [0] ^2 - 14 * x [1] + 6 * x [0] * x [1] + 3 * x [1] ^2))
    term2 = (30 + (2 * x [0] - 3 * x [1]) ^2 * (18 - 32 * x [0] + 12 * x [0] ^2 + 48 * x [1] - 36 * x [0] * x [1] + 27 * x [1] ^2))
    return term1 * term2
```

### 3.4.3 Forrester function

$$f_{forrester}(x) = (6x - 2)^2 sin(12x - 4)$$

Table 33. Pseudo Code Forrester function

```
function Forrester(x):
    term1 = (6 * x [0] - 2) ^2 * sin (12 * x [0] - 4)
    term2 = (6 * x [1] - 2) ^2 * sin (12 * x [1] - 4)
    return term1 + term2
```

It is evaluated on $x \in [0, 1]$ (Forrester et al. 2008)

### 3.4.4 DeVilliersGlasser02

Gavana (2016 and subsequently Layeb (2022) found that DeVilliersGlasser02 is harder to solve than others.

$$f_{devilliersglasser02}(x) = (2x_1 - 3x_2)^2 + 18x_1 - 32x_2 + 12x_1^2 + 48x_2 + 27x_2^2$$

Here, $t_i = 0.1(1 - i)$ and $y_i = 53.81(1.27^{t_i})tanh(3.012t_i + sin(2.13t_i))cos(e^{0.507}t_i)$

Table 34. Pseudo code DevilliersGlasser02

```
function DeVilliersGlasser2(x):
   n = length(x)
   t = [0.1 * (i - 1) for i in range (1, 25)]
   y = [53.81 * (1.27^t) * tanh (3.012 * t + sin (2.13 * t)) * cos (exp (0.507) * t) for t in t]

   if length(x)! = 5:
      raise ValueError ("Input vector x must have 5 dimensions.")

   result = 0.0
   for i in range (24):
      term = (
         x [0] * x [1] ^t[i] * tanh (x [2] * t[i] + sin (x [3] * t[i])) * cos(t[i] * exp (x [4])) - y[i]
      )
      result += term^2

   return result
```

The function is evaluated on $x \in [1, 60]\ for\ 1, \ldots, n$. Here the global optimal $f_{devilliersglasser02}(x^*) = 0$.

Table 31 presents the report. Here too, ADEDS outperforms DE in most of the categories except Beale where although the global minima are achieved by both DE and ADEDS, the output is not statistically significant. It suggests that the observed variations in performance between the two groups (in this case, ADEDS and DE) could be attributable to random variability or by chance rather than a systematic and meaningful difference.

Table 35. ADEDS with other benchmark test report

| Sl no | Benchmark function | | Mean fitness | Std dev | t-stats | p-value |
|---|---|---|---|---|---|---|
| 1 | Beale | DE | 0.000 | 0.000 | 1.000 | 0.330 |
| | | ADEDS | 0.000 | 0.000 | | |
| 2 | Goldstein price | DE | 44.868 | 38.521 | -3.260 | 0.000** |
| | | ADEDS | 2.990 | 0.000 | | |
| 3 | Forrester | DE | -44.183 | 17.378 | 5.548 | 0.017*** |
| | | ADEDS | -12.041 | 0.000 | | |
| 4 | DeVilliersGlasser02 | DE | 143.644 | 201.111 | -6.576 | 0.000*** |
| | | ADEDS | 47.938 | 142.900 | | |

## 3.5 Practical implications

To this end, we see the all-inclusive benchmarking tests on a diverse set of benchmark functions showcasing ADEDS's versatility and effectiveness. ADEDS consistently outperforms traditional DE in terms of convergence speed and solution quality across a range of optimization challenges, including functions with many local optima, plate-shaped, valley-shaped, stretched-shaped, and noisy functions. For most of the vital test problems, such as Ackley, Matyas, Booth, Goldstein-Price, Beale, Bukin, Levy, McCormic, Six-Hump Camel, and Three-Hump Camel, the algorithm did exceedingly well. This gives us confidence that ADEDS is appropriate for supply chain analytics and capable of addressing the complexities and challenges of supply chain management. Its adaptability, performance, and ability to handle various optimization landscapes make it a relevant and promising approach for improving supply chain operations and decision-making. ADEDS can be used to address the challenges by finding solutions that optimize various supply chain parameters, such as inventory levels, production schedules, routing plans, and supplier selections.

While ADEDS designed and evaluated in the context of single-objective optimization it can potentially be adapted to handle multi-objective optimization problems by incorporating suitable mechanisms or modifications to address multiple conflicting objectives simultaneously.

## Conclusion

The Adaptive Differential Evolution with Diversification Strategies (ADEDS) algorithm is a population-based method for solving complex real-parameter single-objective optimization problems. ADEDS is shown to excel in handling non-convex optimization problems, where traditional DE struggles to escape local optima. Visualizations and analysis demonstrated its capability to explore and adapt to complex landscapes. ADEDS dynamically modifies critical parameters based on the optimization process conditions and includes diversification measures to avoid early convergence and encourage solution space exploration. The study analyzed ADEDS performance across various benchmark functions, where the algorithm achieved good success rates in various settings, making it a promising tool for real-world optimization challenges. This study contributes to the growing body of information on optimization algorithms and their application in real-world problem solving.

# Appendix 1

Table 36. Success rate for various test functions and for numbers of iterations.

| Sl no | Functions | Number of runs | | | | | | | | | | | |
|---|---|---|---|---|---|---|---|---|---|---|---|---|---|
| | | 10 | | 50 | | 100 | | 200 | | 300 | | 500 | |
| | | ADEDS | DE | ADEDS | DE | ADEDS | DE | ADEDS | DE | ADEDS | DE | ADEDS | DE |
| 1 | Rastrigin | 0.80 | 0.00 | 100.0 | 0.00 | 0.94 | 0.00 | 0.98 | 0.00 | 0.99 | 0.00 | 1.00 | 0.00 |
| 2 | Ackley | 0.70 | 0.00 | 0.67 | 0.00 | 0.89 | 0.00 | 0.99 | 0.00 | 0.94 | 0.00 | 0.93 | 0.00 |
| 3 | Cross-in-tray | 0.82 | 0.00 | 0.78 | 0.00 | 0.89 | 0.00 | 0.98 | 0.00 | 0.99 | 0.00 | 1.00 | 0.00 |
| 4 | Goldstein-Price | 0.68 | 0.00 | 0.56 | 0.00 | 0.79 | 0.00 | 0.87 | 0.00 | 0.98 | 0.00 | 0.97 | 0.00 |
| 5 | Egg-holder | 0.99 | 0.00 | 1.00 | 0.00 | 0.98 | 0.99 | 0.98 | 0.00 | 0.99 | 0.00 | 0.99 | 0.00 |
| 6 | Levy | 0.79 | 0.00 | 0.89 | 0.00 | 0.95 | 0.00 | 0.98 | 0.00 | 0.99 | 0.00 | 1.00 | 0.00 |
| 7 | Drop-wave | 0.91 | 0.00 | 0.89 | 0.00 | 0.95 | 0.00 | 0.98 | 0.00 | 0.99 | 0.00 | 1.00 | 0.00 |
| 8 | Schwefel | 0.55 | 0.00 | 0.39 | 0.00 | 0.68 | 0.00 | 0.69 | 0.00 | 0.70 | 0.00 | 0.70 | 0.00 |
| 9 | Bukin | 0.89 | 0.00 | 0.89 | 0.00 | 0.95 | 0.00 | 0.98 | 0.00 | 0.99 | 0.00 | 1.00 | 0.00 |
| 10 | Shubert | 0.00 | 0.00 | 0.00 | 0.00 | 0.00 | 0.00 | 0.00 | 0.00 | 0.00 | 0.00 | 0.00 | 0.00 |
| 11 | Schaffer | 0.80 | 0.00 | 0.89 | 0.00 | 0.95 | 0.00 | 0.98 | 0.00 | 0.99 | 0.00 | 1.00 | 0.00 |
| 12 | McCormick | 0.79 | 0.00 | 0.83 | 0.00 | 0.98 | 0.00 | 0.99 | 0.00 | 0.98 | 0.00 | 0.97 | 0.00 |
| 13 | Booth | 0.80 | 0.0 | 0.90 | 0.00 | 0.89 | 0.00 | 0.93 | 0.00 | 1.00 | 0.00 | 1.00 | 0.00 |
| 14 | Matyas | 0.67 | 0.00 | 0.81 | 0.00 | 0.91 | 0.00 | 0.98 | 0.00 | 1.00 | 0.00 | 1.00 | 0.00 |
| 15 | Rosenbrock | 0.80 | 0.00 | 0.64 | 0.00 | 0.0.69 | 0.00 | 0.56 | 0.00 | 0.70 | 0.00 | 1.00 | 0.00 |
| 16 | Three-hump camel | 0.75 | 0.00 | 0.77 | 0.00 | 0.75 | 0.00 | 0.88 | 0.00 | 0.84 | 0.00 | 0.89 | 0.00 |
| 17 | Six-hump camel | 0.72 | 0.00 | 0.69 | 0.00 | 0.87 | 0.00 | 0.98 | 0.00 | 0.99 | 0.00 | 0.97 | 0.00 |
| 18 | Dixon price | 0.90 | 0.00 | 0.89 | 0.00 | 0.95 | 0.00 | 0.98 | 0.00 | 0.99 | 0.00 | 1.00 | 0.00 |
| 19 | DeVilliersGlasser02 | 0.00 | 0.00 | 0.23 | 0.00 | 0.28 | 0.00 | 0.58 | 0.00 | 0.61 | 0.00 | 0.60 | 0.00 |
| 20 | Himmelblau | 0.00 | 0.00 | 0.00 | 0.00 | 0.00 | 0.00 | 0.00 | 0.00 | 0.00 | 0.00 | 0.00 | 0.00 |
| 21 | Forrester | 0.68 | 0.00 | 0.70 | 0.00 | 0.70 | 0.00 | 0.75 | 0.00 | 0.79 | 0.00 | 0.75 | 0.00 |
| 22 | Beale | 0.00 | 0.00 | 0.00 | 0.00 | 0.00 | 0.00 | 0.00 | 0.00 | 0.00 | 0.00 | 0.00 | 0.00 |